\documentclass[10pt,twocolumn,letterpaper]{article}

\usepackage{cvpr}
\usepackage{times}
\usepackage{epsfig}
\usepackage{graphicx}
\usepackage{amsmath}
\usepackage{amssymb}

\usepackage[breaklinks=true,bookmarks=false]{hyperref}

\cvprfinalcopy 

\def\cvprPaperID{5671} 

\begin{document}
	
	\title{Unity Style Transfer for Person Re-Identification}
	
	\author{Chong Liu \textsuperscript{1,2} \and Xiaojun Chang \textsuperscript{3} \and Yi-Dong Shen \textsuperscript{1}\thanks{Corresponding author} \and
		\textsuperscript{1} State Key Laboratory of Computer Science, Institute of Software, Chinese Academy of Sciences, China \\
		\textsuperscript{2} University of Chinese Academy of Sciences, Beijing 100049, China \\
		\textsuperscript{3} Monash University, Melbourne, Australia\\
		{\tt\small liuchong@ios.ac.cn, cxj273@gmail.com, ydshen@ios.ac.cn}
	}
	
	\maketitle

	\begin{abstract}
		Style variation has been a major challenge for person re-identification, which aims to match the same pedestrians across different cameras. Existing works attempted to address this problem with camera-invariant descriptor subspace learning. However, there will be more image artifacts when the difference between the images taken by different cameras is larger. To solve this problem, we propose a UnityStyle adaption method, which can smooth the style disparities within the same camera and across different cameras. Specifically, we firstly create UnityGAN to learn the style changes between cameras, producing shape-stable style-unity images for each camera, which is called UnityStyle images. Meanwhile, we use UnityStyle images to eliminate style differences between different images, which makes a better match between query and gallery. Then, we apply the proposed method to Re-ID models, expecting to obtain more style-robust depth features for querying. We conduct extensive experiments on widely used benchmark datasets to evaluate the performance of the proposed framework, the results of which confirm the superiority of the proposed model.
	\end{abstract}
	
	\section{Introduction}
	
	Person re-identification (re-ID) \cite{IDE} is a multi-camera query task. Given the image of one or a group of target people, the same person is identified from the database of multi-camera. There are multiple cameras in the re-id task, and the images style taken by each camera are often different for the same person due to various factors (environment, light, \textit{etc.}). Besides, even the same camera will take different styles of images due to the different time (morning, noon, afternoon, \textit{etc.}). Therefore, the style of the image changes, which has a considerable impact on the final task results.
	
	\begin{figure}[t]
		\vspace{-0.4 em}
		\begin{center}
			\includegraphics[width=1\linewidth]{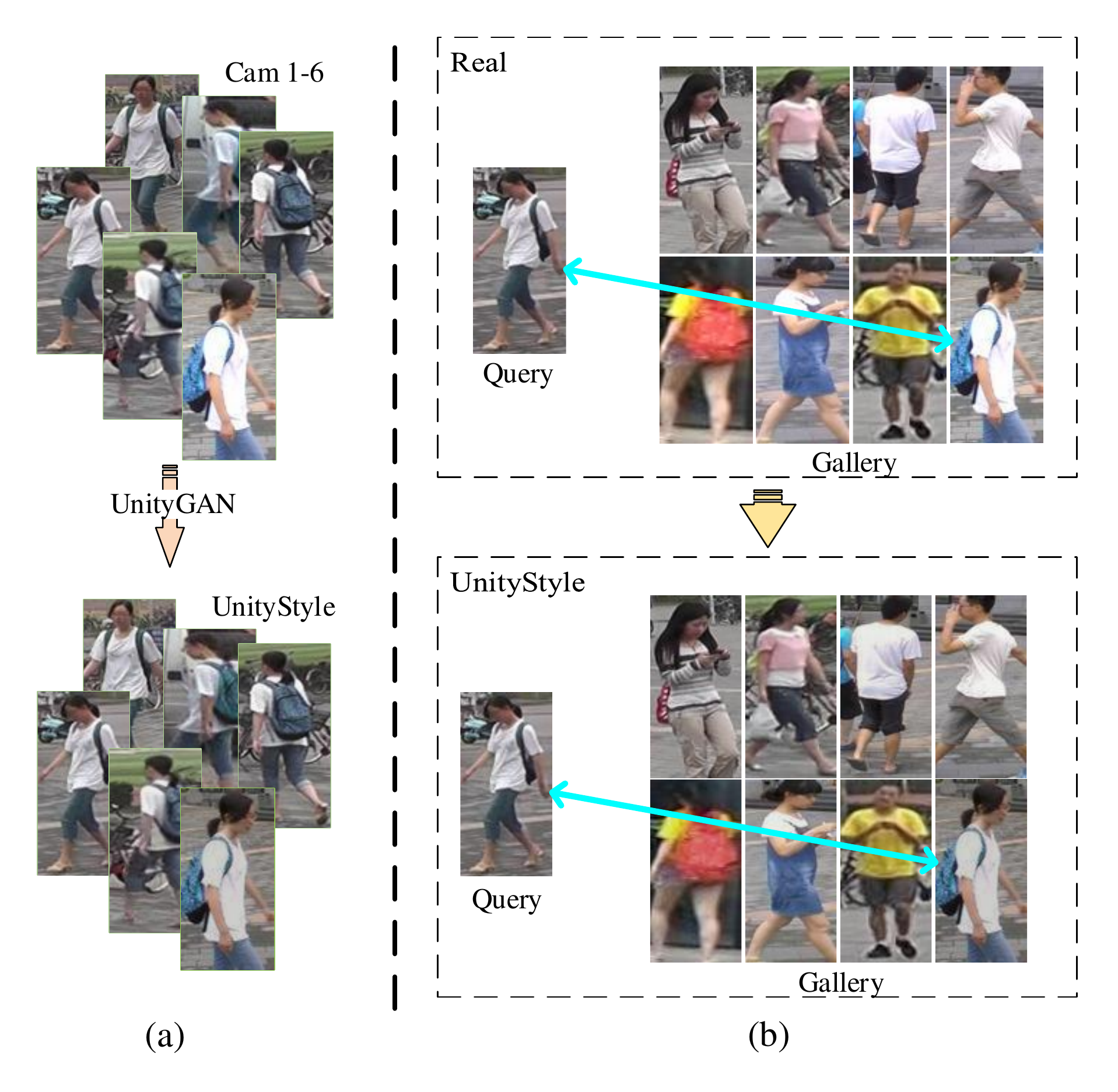}
		\end{center}
		\vspace{-1.4 em}
		\caption{\textbf{(a)} Different cameras often have significant style differences, and UnityGAN is used to convert image styles to each other to achieve a unified style. \textbf{(b)} By using UnityStyle, the query collection is similar to the gallery collection image style, making it easier to find the right match.}
		\label{fig:introduction}
	\end{figure}

	There are some methods available to address different camera style problems. One solution is to obtain stable feature representations between different cameras. There are KISSME\cite{KISSME}, DNS \cite{DNS} and so on in the traditional method. IDE \cite{IDE}, PCB \cite{PCB} solve problems through deep representation learning. Another way is to use the GANs \cite{CycleGAN} to learn the style disparities between different cameras and to enhance the data through the style transfer method, resulting in CamStyle \cite{CamStyle}.
	
\begin{figure}[t]
		\vspace{-0.4 em}
		\begin{center}
			\includegraphics[width=1\linewidth]{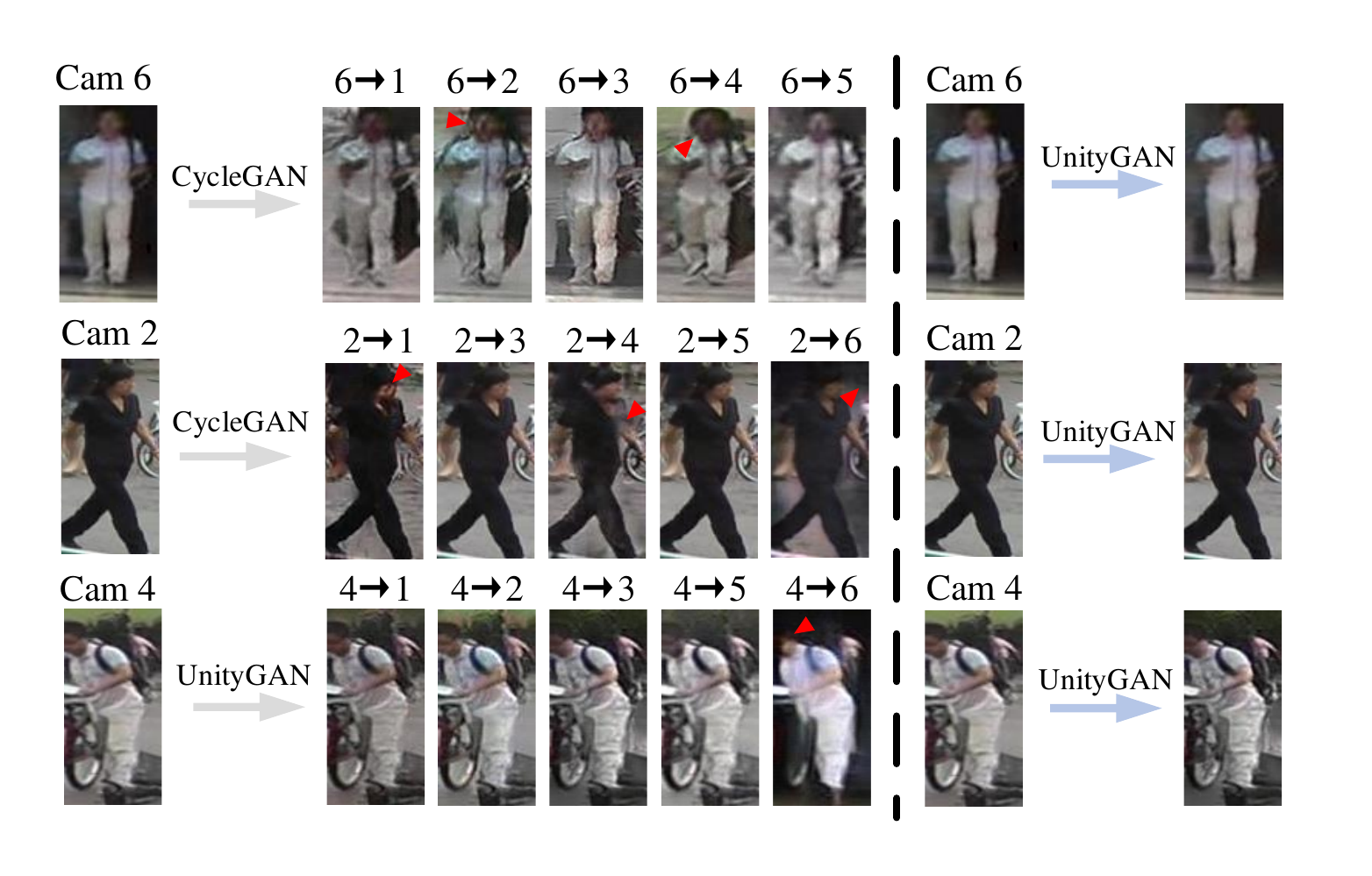}
		\end{center}
		\vspace{-1.4 em}
		\caption{Examples of generated by CycleGAN and UnityGAN in Market-1501. UnityGAN can produce better style-transferred images. As can be seen from the comparison in the figure, UnityGAN solves the problem that CycleGAN generates error images, and no longer has the problem of marking.}
		\label{fig:GANsVS}
	\end{figure}

	Compared with the previous methods, this paper proposes an alternative solution to smooth the style disparities across different cameras and within the same camera. We start from CamStyle and find that although this method can learn and transform styles between different cameras, there are still some problems.
	\textbf{1)} There will be image artifacts in the transfer sample generated by CycleGAN \cite{CycleGAN}, especially for the shaped part, which produces a considerable number of error images (Fig. \ref{fig:GANsVS}). 
	\textbf{2)} The generated enhanced images introduce some noise to the system, and the Label Smooth Regularization (LSR) is required to adjust the network performance.
	\textbf{3)} The generated enhanced images can only be used as a data enhancement method to extend the training set, which is not effective.
	\textbf{4)} The number of models that need to be trained is $C_C^2$ \textbf{(where $C$ is the number of cameras)}, which means that as the number of cameras increases, the number of models that need to be trained will become larger and larger, which is not applicable in scenarios where computing resources are insufficient. 
	
	To access the above problems, we build a UnityStyle adaption method, which can smooth the style disparities within the same camera and across different cameras. We overcome the problem of CycleGAN's easy deformation with UnityGAN and generate style-unity images with higher quality (Fig. \ref{fig:GANsVS}). Further, we rely on the style data of each camera learned by UnityGAN to get a UnityStyle image suitable for all camera styles, which makes the generated enhanced images more efficient. Finally, we combine real images and UnityStyle images as the new data augmentation training set.
	
	The UnityStyle adaption method has the following advantages. \textbf{First}, as a data enhancement scheme, the generated enhanced sample can be treated the same as original images. Thus, LSR is no longer needed for UnityStyle images. \textbf{Secondly}, by adapting to different camera styles, it is also robust to style changes within the same camera. \textbf{Thirdly}, it does not require additional information, and all enhancements are derived from the general information of the re-ID task. \textbf{Finally}, it only needs to train $C$ UnityGAN models, which only require a small number of computing resources.
	
	To summarize, this paper has the following contributions: 
	\begin{itemize}
		\item The UnityStyle method for Re-ID is proposed to generate shape-stable style-changing enhanced images, which can be treated equally as the real images. LSR is no longer needed for the style transferred images.
		\item The UnityStyle does not require training a large number of models. It only needs to train $C$ UnityGAN models, which use a small number of computing resources.
		\item We propose a novel data enhancement scheme. We use UnityStyle to eliminate style differences between different images, which makes a better match between the images of query and gallery, and the generated enhanced images more efficient.
		\item Our experiments show that UnityStyle can be easily applied to a wide range of models and can significantly improve experimental performance.
	\end{itemize}	
	
	\section{Related Work}
	
	\subsection{Deep learning person re-identification}
	In the field of person re-ID, many deep learning methods \cite{IDE,DML1,DML2,DML3,attention,DML4,DML5} have been proposed. \cite{IDE} uses the ImageNet \cite{ImageNet} pre-trained model, and trains the model as image classification in re-ID, which is called the ID-discriminative embedding (IDE). Problems with overfitting and insufficient sample types have emerged as the CNN models become sufficiently complex. There are some data augmentation methods been proposed to solve this problem. \cite{DA2} used DC-GAN \cite{DC-GAN} to improve the discriminative ability of learned CNN embeddings. \cite{FD-GAN,PN-Reid,JD-Reid} focus on the person’s pose, and improve the final performance by generating different pose images.  More related to this work, \cite{CamStyle} proposed CamStyle data augmentation approach which transfers images from one camera to the style of another camera. 
	
	\subsection{Generative Adversarial Networks}
	Generative adversarial networks (GANs)\cite{GANs} have achieved remarkable success since its introduction. Recently, GANs are used in image translation \cite{IT1,DiscoGAN,CycleGAN}, style transfer \cite{ST1,ST2} and image editing \cite{IE1}. In \cite{ST1}, a style transfer model was proposed by extracting and reconstructing the substance and style of the image. \cite{IT1} showed through Pix2Pix that GANs can learn style mapping between different domains. Recently, the Pix2Pix-like framework have evolved to apply to unsupervised pairs \cite{DiscoGAN,CycleGAN}. \cite{GANimorph} proposed a new method built upon the DiscoGAN \cite{DiscoGAN} and CycleGAN \cite{CycleGAN} architectures, which overcome the limitations of shape changes through more effective learning.
	
	\subsection{IBN-Net}
	
	According to \cite{IBNNET}, Batch Normalization (BN) \cite{BN} improves the sensitivity of features to image content, and Instance Normalization (IN) \cite{IN}  improves robustness to style class changes. Therefore, IN achieves better results than BN in the field of style migration. Shallow features are related to style, and deep features are related to high-level features (content features such as face and gesture). This paper raises two criteria:
	
	\begin{itemize}
		\item IN is only added to the shallow network and is not added to the deep network. Because the feature of IN extraction reduces the difference between images, it cannot be placed in the deep layer to influence the classification effect.
		\item The shallow BN layer should also be retained to ensure that content-related information can be smoothly passed into the deep layer.
	\end{itemize}
	
	\begin{figure}[t]
		\vspace{-0.4 em}
		\begin{center}
			\includegraphics[width=1\linewidth]{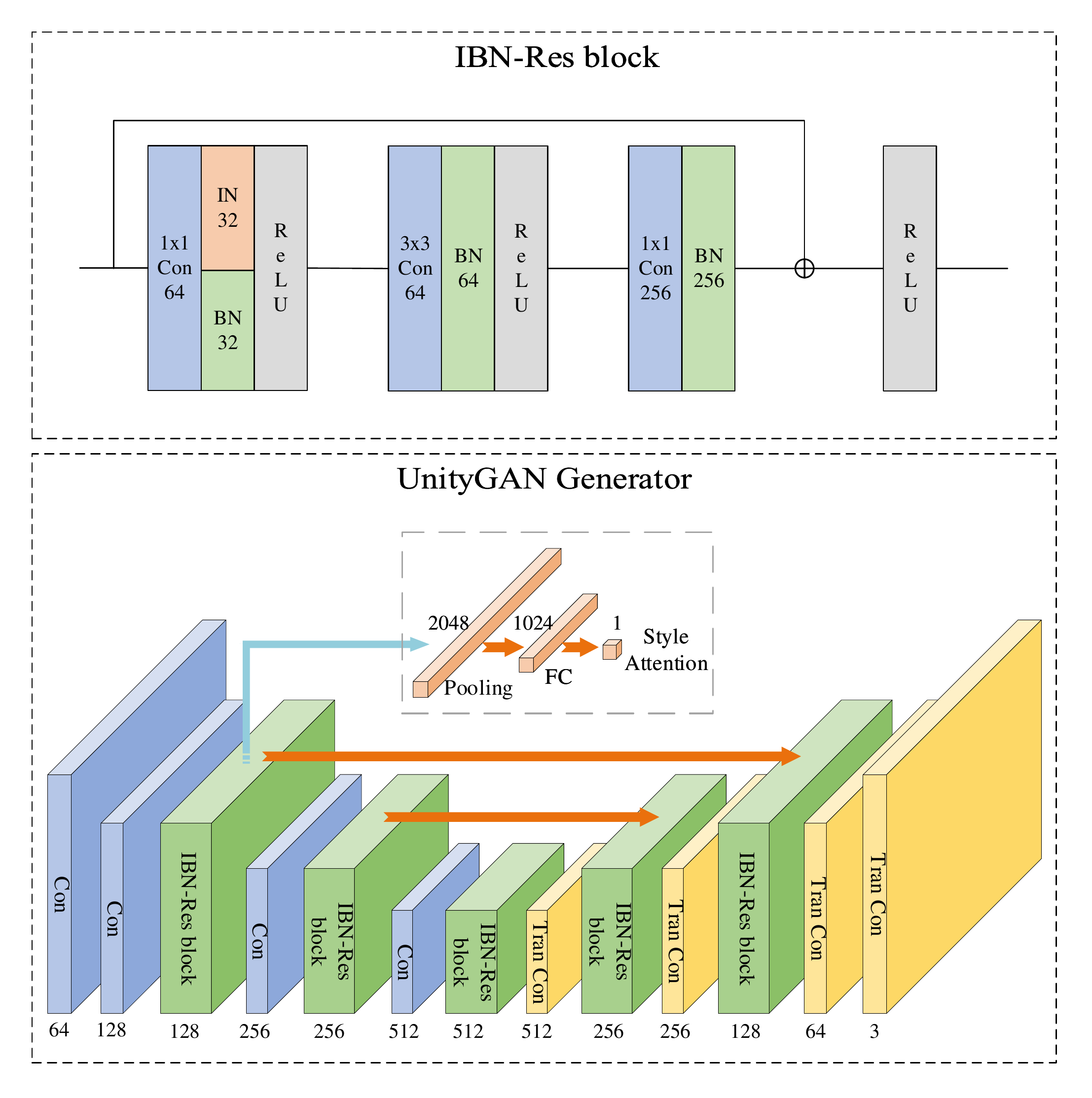}
		\end{center}
		\vspace{-1.4 em}
		\caption{Generators from different unsupervised image translation models. \textbf{Blue box} is convolution layer.
			\textbf{Green box} is residual block.
			\textbf{Yellow box} is transposed convolution layer.
			\textbf{Orange box} is Style Attention module.
		}
		\label{fig:GANimorph}
	\end{figure}
	\section{The Proposed Method}
	
	In this section, we will explain in detail the method we proposed. First, we create UnityGAN and make it suitable for re-ID task. Then, we propose how to generate UnityStyle images. Finally, our proposed method is shown to be applied to multiple models and explain the detailed operation of the pipeline.
	
	\subsection{UnityGAN}
	\begin{figure}[t]
		\vspace{-0.4 em}
		\begin{center}
			\includegraphics[width=1\linewidth]{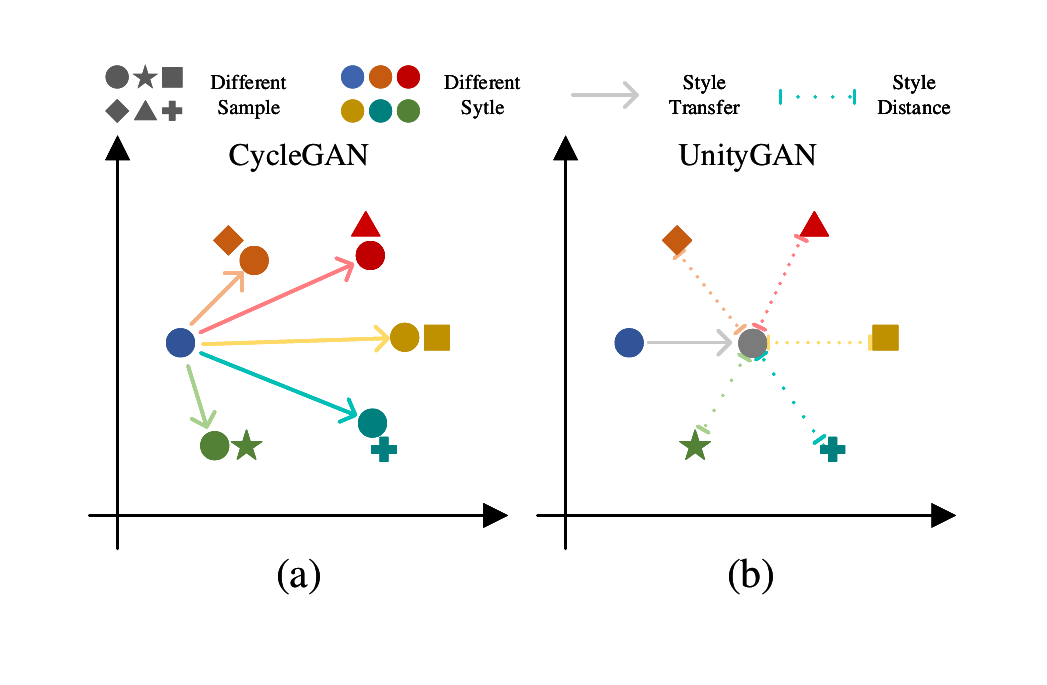}
		\end{center}
		\vspace{-1.4 em}
		\caption{The position of the sample in the style space. \textbf{(a)} Each sample requires multiple style transfer to generate different style samples. \textbf{(b)} Each sample only needs one style transfer to generate a uniform style sample.}
		\label{fig:StyleDis}
	\end{figure}
	In this section, we use UnityGAN for Style Transfer and train the Style Transfer model. UnityGAN absorbs the advantages of DiscoGAN and CycleGAN and improves them. DiscoGAN uses the standard architecture, and its narrow bottleneck layer may prevent the output image from retaining visual details in the input image. CycleGAN introduces residual blocks to increase the capacity of DiscoGAN, but the use of residual blocks on a single-scale layer does not preserve information on multiple scales. UnityGAN combines two networks (Fig. \ref{fig:GANimorph}, bottom), and introduces residual blocks and skip connections in the multi-scale layer, which can retain multi-scale information at the same time, making the transformation more precise and accurate. Through the transfer of multi-scale information, UnityGAN can generate structurally stable images and avoid generating images of the wrong structure (Fig. \ref{fig:GANsVS}, right). Unlike CycleGAN, UnityGAN tries to generate a picture that blends all styles without having to learn a transfer for each style (Fig. \ref{fig:StyleDis}).
	
	Further, we created an IBN-Res block (Fig. \ref{fig:GANimorph}, upper) based on \cite{IBNNET} discussion, which can increase the robustness of style changes while maintaining structural information. Add the IBN-Res block to the UnityGAN model to adapt the model to the style changes and ensure that the model generates a uniform style of fake images.
	
	Given image domains $X$ and $Y$, let $G : X \rightarrow Y$ and $F : Y \rightarrow X$. $D_X$ and $D_Y$ denote discriminators for $G$ and $F$ respectively. To preserve the feature information of the image while changing the style, we add the identity mapping loss\cite{CycleGAN} to the formula. The identity mapping loss can be expressed as:
	
	\begin{equation}
	\begin{split}
	\mathcal{L}_{ID}=&\mathbb{E}_{x\sim l_x}(\|F(x)-x\|_1)\\
	+&\mathbb{E}_{y\sim l_y}(\|G(y)-y\|_1),
	\label{eq:1}
	\end{split}
	\end{equation}
	
	Therefore, UnityGAN's loss function comprises four types of loss normalized terms: the standard GANs loss, the feature matching loss, the identity mapping loss and the cyclic reconstruction loss.
	
	Our total objective function is:
	\begin{equation}
	\begin{split}
	\mathcal{L}_{UnityGAN}(x,y)&=\lambda_{GAN}SLN(\mathcal{L}_{GAN})\\
	+&\lambda_{FM}SLN(\mathcal{L}_{FM})\\
	+&\lambda_{ID}SLN(\mathcal{L}_{ID})\\
	+&\lambda_{CYC}SLN(\lambda_{SS}\mathcal{L}_{SS}+\lambda_{L1}\mathcal{L}_{L1}),
	\label{eq:2}
	\end{split}
	\end{equation}
	Where, $\mathcal{L}_{FM}=\mathcal{L}_{FM_X}(G,D_X)+\mathcal{L}_{FM_Y}(F,D_Y)$, $\mathcal{L}_{GAN}=\mathcal{L}_{GAN_X}(F,D_X,Y,X)+\mathcal{L}_{GAN_Y}(G,D_Y,X,Y)$, $SLN$ is scheduled loss normalization \cite{GANimorph}. With $\lambda_{GAN}+\lambda_{FM}+\lambda_{ID}+\lambda_{CYC}=1$, $\lambda_{SS}+\lambda_{L1}=1$, and all coefficients $\geq 0$.
	
\begin{figure}[t]
		\vspace{-0.4 em}
		\begin{center}
			\includegraphics[width=1\linewidth]{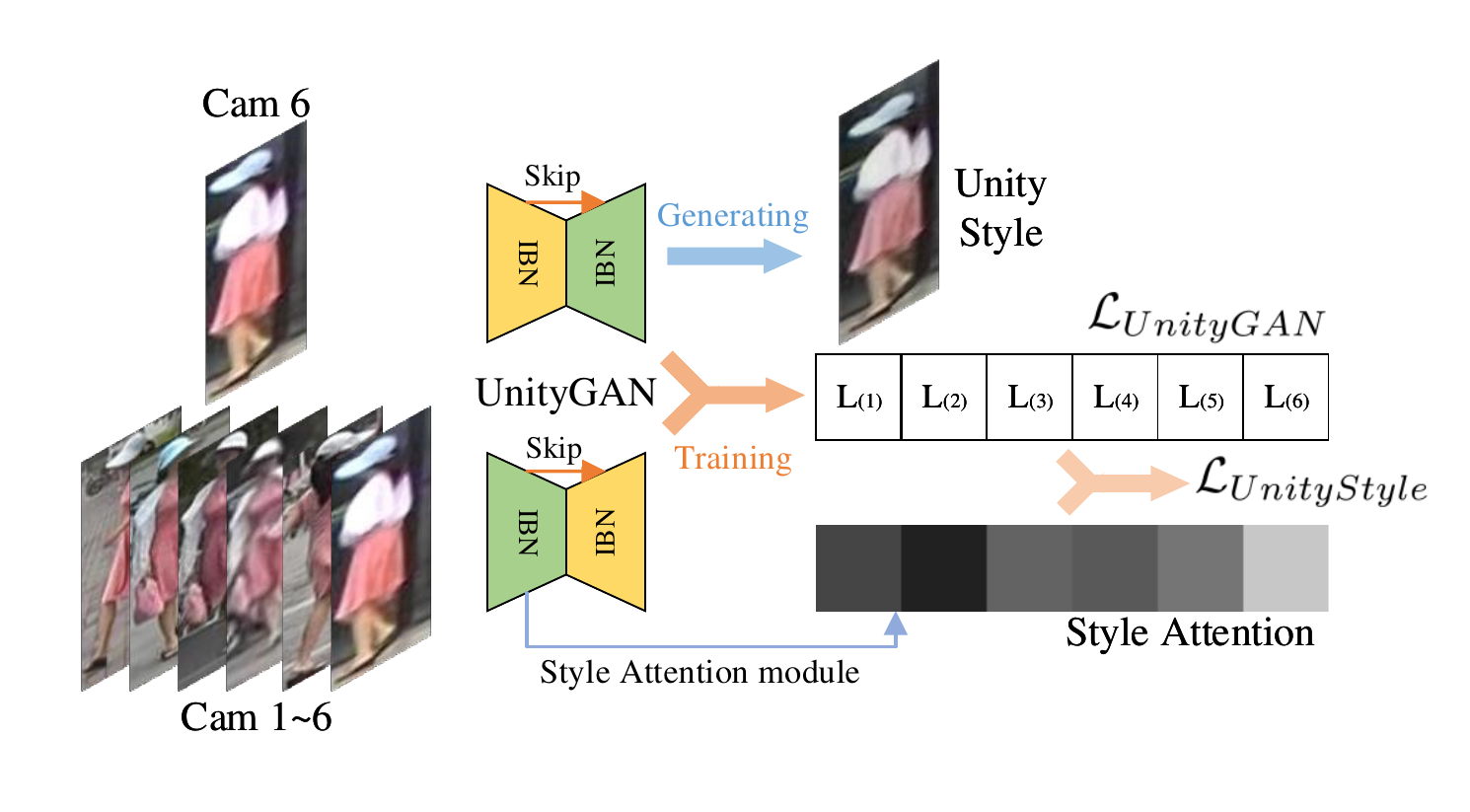}
		\end{center}
		\vspace{-1.4 em}
		\caption{The main process of UnityStyle. In training, UnityGAN uses all camera style pictures for training. In generating, UnityGAN generates UnityStyle for each image.}
		\label{fig:UnityGAN}
	\end{figure}

	In the training phase, we take each camera and all cameras as a group from the training set to train a UnityGAN model (Fig. \ref{fig:UnityGAN}). Same as \cite{CycleGAN} during the training phase, we resize all images to $256\times 256$.
	UnityGAN can generate stable structure pictures and reduce the number of training models. However, images generated by UnityGAN is unstable in style (Fig. \ref{fig:UnityStyle}). To solve this problem, we propose the UnityStyle loss function (Eq. \ref{eq:unity_loss}) in the next section to ensure that UnityGAN generates stable style images.
	
	
	\subsection{UnityStyle}
	
	In this section, we propose the concept of UnityStyle. UnityStyle images are generated from UnityGAN, which can smooth the style disparities within the same camera and across different cameras. Then we use UnityStyle images for model training and prediction to improve the performance of the model.
	
	To ensure that UnityGAN can generate UnityStyle images, we add the Style Attention module to the UnityGAN Generator (Fig. \ref{fig:GANimorph}). Low-level image features get style-related attention features through this module. We define the Style Attention of the input image $x$,
	
	\begin{equation}
	\mathcal{A}(x)=Sigmoid(A_{style}(G_1(x))),
	\label{eq:attention}
	\end{equation}
	where, $A_{style}$ is Style Attention Module, $G_1$ is the first IBN-Res block output of UnityGAN Generator.
	
\begin{figure}[t]
		\vspace{-0.4 em}
		\begin{center}
			\includegraphics[width=1\linewidth]{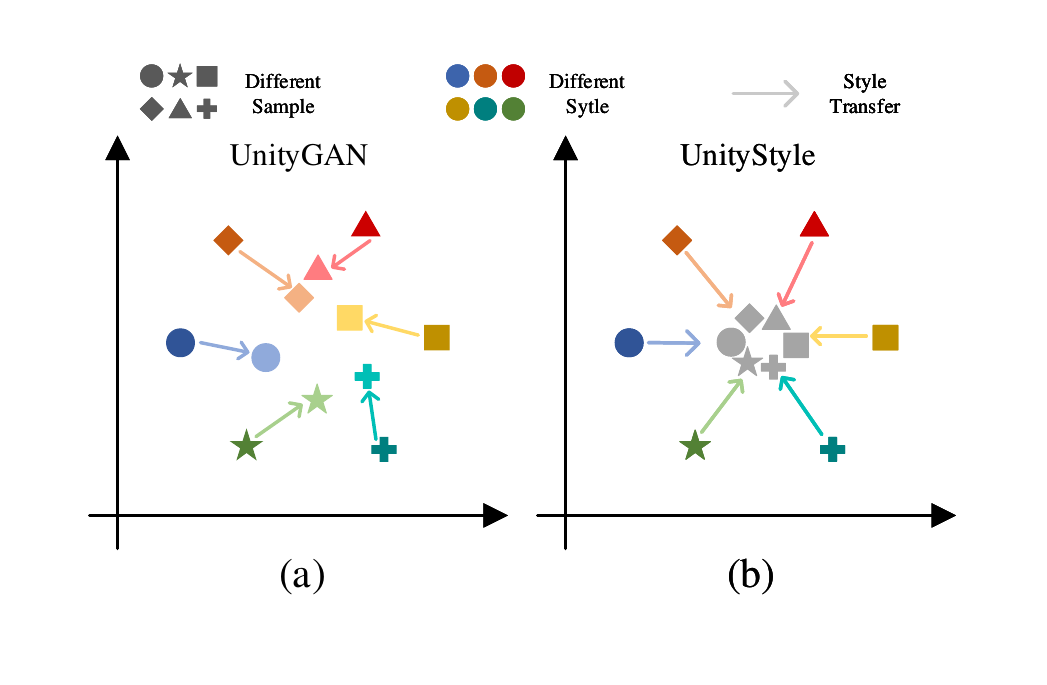}
		\end{center}
		\vspace{-1.4 em}
		\caption{The position of the sample in the style space. \textbf{(a)} Sample can not achieve the complete unity of style after the transfer of UnityGAN without the style attention module. \textbf{(b)} Sample can be completely unified by the transfer of UnityStyle.}
		\label{fig:StyleDisS}
	\end{figure}

	Further, we get the final UnityStyle loss function,
	
	\begin{equation}
	\mathcal{L}_{UnityStyle}=\sum_{c=1}^{L}(\mathcal{A}(y_i^{(c)})\mathcal{L}_{UnityGAN}(x_i,y_i^{(c)})),
	\label{eq:unity_loss}
	\end{equation}
	where, $c$ is camera number, $C$ is the number of cameras, $\mathcal{A}(y_i^{(c)})$ is the $i$th camera's style attention. Under the definition of Eq. \ref{eq:unity_loss}, the model will generate a style -stable picture, and the style of this picture is between all camera styles (Fig. \ref{fig:StyleDisS}).
	
	From Fig. \ref{fig:UnityStyle}, we can see that UnityGAN with the above module can produce style-stable pictures, compared to UnityGAN without Attention. The image generated by UnityGAN under the definition of Eq. \ref{eq:attention} is a uniform style image, and the different style features of multiple cameras are smoothed out. The six images under the six cameras, whose image styles produced by UnityStyle Transfers are no different. Intuitively, our method has better performance, and we will further compare it in the experiment. 
	
	Although using UnityGAN guarantees the stability of the structure, the style and lighting of each image are different (Fig \ref{fig:UnityStyle} middle line). There are multiple style images in the training set, but current GAN methods cannot converge to a unified style (Different styles of images have different losses). The Style Attention module quantifies the image style and introduces $\mathcal{L}_{UnityStyle}$ for training, ensuring the uniformity of the style of the generated images (Fig \ref{fig:UnityStyle} last line). In Fig \ref{fig:UnityGAN}, the lighter the Attention square color of the image, the farther the unity-style is away from the corresponding image style (The loss has a lower weight). It is ensured that images with a large difference in unity-style have a small effect on the style of the generated image, and we can generate images with unity-style. In addition, the style information is mainly contained in the shallow layer. This is why we add Style Attention module in the shallow layer.
	
	Past data enhancements are often performed in test data to ensure that the trained model is more robust, but this is only one aspect of the work. In re-ID task, test set consists of two sets of query and gallery. The test finds the image belonging to someone from the gallery images by his given query image. Under the task of multiple cameras, in Fig. \ref{fig:introduction}, the query image and the corresponding gallery image are often different styles. Therefore, we can start from the test set and ensure that the training model is more robust while making the multi-camera test set style as close as possible (i.e., close the distribution of the test set, as shown by Fig. \ref{fig:introduction}). 
	
	In training, UnityStyle images are used to augment the training set, which makes the model more adaptive.
	In testing, UnityStyle images are test inputs, which make it easier to match between query and gallery.
	The test set consists of two sets of query and gallery. The test finds the image belonging to someone from the gallery images by his given query image. Under the task of multiple cameras, as shown in the upper of Fig. \ref{fig:introduction} (b), the query image and the corresponding gallery image often have different styles. Therefore, we take UnityStyle images as test inputs, which ensures that the training model is more robust and makes the multi-camera test set style as close as possible (i.e. close the distribution of the test set, as shown by the Fig. \ref{fig:introduction} (b) bottom).
	
	\subsection{Deep Re-ID Model}
	
	\begin{figure*}[t]
		\vspace{-0.4 em}
		\begin{center}
			\includegraphics[width=1\linewidth]{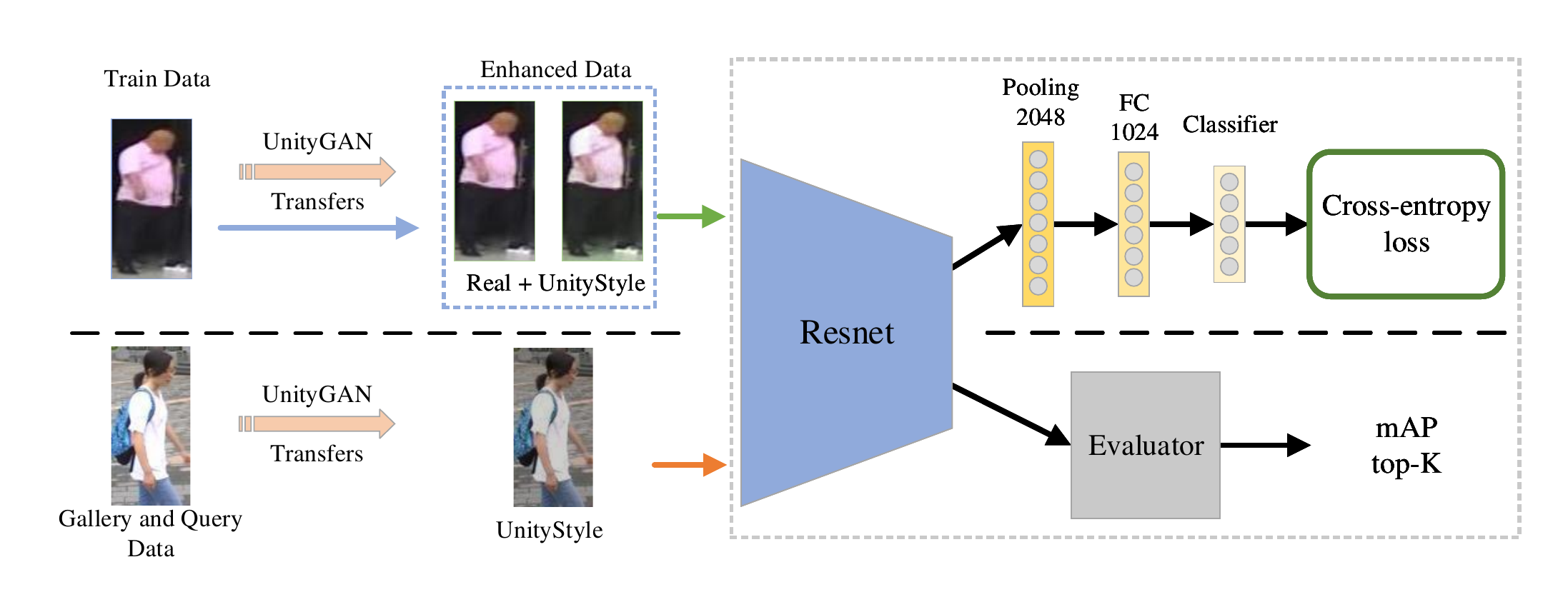}
		\end{center}
		\vspace{-1.4 em}
		\caption{The pipeline of our method (Take IBN as an example, gray boxes can be replaced with other models, such as PCB, st-ReID). In training, the UnityStyle images are generated by training data using the UnityStyle Transfers, and Resnet extracts the features of these two types of data, which will be trained by the classifier. In testing, the UnityStyle images are generated by the gallery and query data through UnityStyle Transfers, and then we use the unified style images to extract features by Resnet, which are evaluated to obtain the results.}
		\label{fig:model}
	\end{figure*}
	
	There are already many excellent deep Re-ID models (such as IDE\cite{IDE}, PCB\cite{PCB} and st-ReID\cite{st-ReID}), and our method can be easily applied to these models. Our approach is based on the ID-discriminative embedding (IDE) \cite{IDE} as an example, and the backbone of the network is ResNet. The basic flow of our proposed pipeline is shown in Fig. \ref{fig:model}. In order to resemble a human body, the size of the input image is $256 \times 128$. In the training phase, we need to ensure that the final classification layer output is consistent with the number of labels in the training set. As shown in Fig. \ref{fig:model}, we replace the last classification layer with two fully connected layers. At the test phase, we use the model's 2048-dimensional feature output for evaluation. Evaluator uses output features to calculate mean average precision(mAP) and top-K represents the proportion of the correct results in the top K retrieval results.
	
	\subsection{Pipeline}
	
	In this section, we will demonstrate the Pipeline of the model from the stages of training and testing in detail (Fig. \ref{fig:model}).
	
	\subsubsection{Training}
	Before training, we used trained UnityGAN Transfers to generate UnityStyle images. We combine real images and UnityStyle images as an enhanced training set. In training, we take the image in the enhanced dataset as input, and all input images sizes are specified as $256\times 128$. We randomly sample $N$ real images and $N$ UnityStyle images. Under the above definition, we get the loss function,
	
	\begin{equation}
	\mathcal{L}_{REID}=\frac{1}{N}\sum_{i=1}^N(\mathcal{L}_R^i+\mathcal{L}_U^i),
	\label{eq:4}
	\end{equation}
	where $\mathcal{L}_R^i=\mathcal{L}_{Cross}(x_R^i)$ and $\mathcal{L}_U^i=\mathcal{L}_{Cross}(x_U^i)$. $x_R^i$ is real image sample and $x_U^i$ is UnityStyle image sample. $\mathcal{L}_{Cross}$ is cross-entropy loss function,
	\begin{equation}
	\mathcal{L}_{Cross}(x)=-\sum_{l=1}^L\log(p(l))q(l),
	\label{eq:5}
	\end{equation}
	where, $L$ is the number of labels, $p(l)$ is the probability that the label of $x$ is predicted to be $l$, $q(l) = \{1 \text{ if } l=y | 0 \text{ if } l\neq y\}$ is the ground-truth distribution. According to the previous description, $\sum_{l=1}^Lp(l)=1$. In $q(l)$,  $y$ is the ground-truth label corresponding to the current image. 
	
	For $x$ with ground-truth label $y$, according to Eq. \ref{eq:5} we can get,
	\begin{equation}
	\mathcal{L}_{Cross}(x)=-\log(p(y)).
	\label{eq:7}
	\end{equation}
	
	Further according to Eq. \ref{eq:7} and Eq. \ref{eq:4}, we can get,
	
	\begin{equation}
	\mathcal{L}_{REID}=-\frac{1}{N}\sum_{i=1}^N\log(p_R^ip_U^i),
	\label{eq:8}
	\end{equation}
	where, $p_R^i$ is the probability that the i-th real image is predicted correctly, $p_U^i$ is the probability that the i-th UnityStyle image is predicted correctly.
	
	As mentioned before, real images and fake images are treated differently in \cite{CamStyle}, using different loss functions for training. In our version, we overcome the shortcomings of \cite{CamStyle}, and the resulting images have the same status as real images. Therefore, we no longer use LSR during training.
	
	\subsubsection{Testing}
	For testing, it is divided into query and gallery data sets. Previously we introduce the concept of UnityStyle to ensure that the images of two datasets generate the corresponding UnityStyle images by UnityStyle Transfers before starting the test. We use the generated UnityStyle images as the new input to test. As discussed before, UnityStyle narrows the gap between the style of elements in the test set (Fig. \ref{fig:introduction}), which can further improve query performance.
	
	\section{Experiments}
	
	\subsection{Datasets and Evaluation Setting}
	We use two common and representative re-ID datasets, Market-1501 \cite{Market} and DukeMTMC-reID \cite{DA2}. Top-1 accuracy and mAP were used to evaluate both datasets.
	
	\textbf{Market-1501}
	\cite{Market} Contains 32,668 labeled images from 6 camera angles of view, which has 1501 identities and some noise images. The data set is divided into three parts: 751 individuals used for training containing 12936 images, 750 individuals for testing containing 19732images, and 3368 images for the query in total.
	
	\textbf{DukeMTMC-reID}
	\cite{DA2} Contains 36,411 labeled images from 8 camera views, which has 1404 identities and some noise images. Similar to Market-1501, the data set is divided into three parts: 702 individuals used for training containing 16522 images, 702 individuals for testing containing 17661 images, and 2228 images for the query in total.
	
	\textbf{Evaluation metrics}
	\cite{Market} The Cumulative Matching Characteristic (CMC) and mAP are used to evaluate the person re-ID task. For each query, its average precision is calculated from the accuracy recall curve. Then the average of the average precision in the query is measured using the mAP. The retrieval accuracy is reflected in the CMC, and the recall rate is reflected in the mAP.
	
	\subsection{Experiment Settings}
	
	\subsubsection{UnityStyle Transfer Model}
	For the style transfer part, the hyperparameter is introduced in Eq. \ref{eq:2}, setting $\lambda_{GAN}=0.25$, $\lambda_{FM}=0.1$, $\lambda_{ID}=0.15$, $\lambda_{CYC}=0.5$, $\lambda_{SS}=0.7$, $\lambda_{L1}=0.3$ for all datasets. Before the start of the follow-up experiment, we need to train a style transfer for each camera, and we need to train multiple style transfers under different camera numbers. So we need to train 6 transfers for Market-1501, and 8 transfers for DukeMTMC-reID. Each style transfer uses $256 \times 256$ images for 50 epochs training, making the resulting image to an excellent level.
	
	\subsubsection{Deep Re-ID Model}
	Our approach can be applied to many models, this paper uses three models (IDE\cite{IDE}, PCB\cite{PCB}, st-ReID\cite{st-ReID}) to verify the effectiveness of our approach. We trained using the baseline model with an input image size of $256\times 128$ and processed training images using random cropping \cite{ImageNet}, random horizontal flipping \cite{RF} and random erasing \cite{RE} during training. We performed 50 epochs training on the model, using 128 real images and 128 UnityStyle images for each batch of each epoch, and using the learning rate = $0.1$ for SGD to solve the model. 
	
	\subsection{Evaluation}
	We compare the proposed method with existing methods on the Market-1501 and DuckMTMC-ReID (Table \ref{tab:market}, \ref{tab:duck}). To verify the broad applicability of UnityStyle, we use IDE, PCB and st-ReID as the baseline. Then we apply our method to the baseline to verify the validity. The results of experiments show that our method achieves the state of the arts on both data sets.
	
	In Market-1501, our method has varying degrees of improvement compared to the baseline. Compared with CamSyle, we also considered similar camera styles and achieved more effective data enhancement. Further we used re-ranking \cite{reranking} technology to make our final experimental results reach\textbf{ top-1 = 98.5\%} and \textbf{mAP = 95.8\%} with st-ReID. Our method achieved significant improvements over the underlying method. From the results we can get, our basic method not only improves the accuracy of top-1, but also ensures the accuracy of candidate results, so the final result is significantly improved after the introduction of re-ranking technology.
	
	Same as Market-1501, our method has achieved good performance in DuckMTMC-ReID. Further, we used re-ranking technology to achieve our final experimental results of \textbf{top-1 = 95.1\%}, \textbf{mAP = 93.6\%} with st-ReID.
	
	\begin{table}
		\small
		\begin{center}
			\setlength{\tabcolsep}{2.7mm}{
				\begin{tabular}{|p{2.7cm}|c c c|c|}
					\hline
					Measure (\%) 	& top-1 & top-5 & top-10 & mAP \\
					\hline\hline
					PAN \cite{PAN} 				& 82.8 & -    & - 	 & 63.4 \\
					GAN \cite{DA2} 				& 83.9 & -    & - 	 & 66.1 \\
					TriNet \cite{TripleNet} 	& 84.9 & 94.2 & - 	 & 69.1 \\
					PAN+RE \cite{PAN} 			& 85.8 & 93.4 & - 	 & 76.6 \\
					CamStyle \cite{CamStyle} 	& 89.5 & - 	  & - 	 & 71.6 \\
					PSE+ECN \cite{PSE} 			& 90.3 & 94.5 & - 	 & 84.0 \\
					HA-CNN \cite{attention} 	& 91.2 & -    & - 	 & 75.7 \\
					\hline
					IDE \cite{IDE} 				& 85.7 & 93.1 & 95.3 & 65.9 \\
					PCB \cite{PCB} 				& 91.2 & 97.0 & 98.2 & 75.8 \\
					st-ReID \cite{st-ReID} 		& 98.0 & 98.9 & 99.1 & 95.5 \\
					\hline
					\textbf{IDE+UnityStyle} 	& 93.2 & 96.1 & 96.9 & 89.3 \\
					\textbf{PCB+UnityStyle} 	& 95.8 & 97.9 & 98.7 & 93.6 \\
					\textbf{st-ReID+UnityStyle} & \textbf{98.5} & \textbf{99.0} & \textbf{99.1} & \textbf{95.8} \\
					\hline
			\end{tabular}}
		\end{center}
		\caption{Evaluation on the Market-1501 dataset.}
		\label{tab:market}
	\end{table}
	
	\begin{table}
		\small
		\begin{center}
			\setlength{\tabcolsep}{2.7mm}{
				\begin{tabular}{|p{2.7cm}|c c c|c|}
					\hline
					Measure (\%) & top-1 & top-5 & top-10 & mAP \\
					\hline\hline
					TriNet \cite{TripleNet} 	& 72.4 & - 	  & - 	 & 53.5 \\
					CamStyle \cite{CamStyle} 	& 78.3 & - 	  & - 	 & 57.6 \\
					PSE+ECN \cite{PSE} 			& 79.8 & 89.7 & 92.2 & 62.0 \\
					HA-CNN \cite{attention} 	& 80.5 & - 	  & - 	 & 63.8 \\
					MLFN \cite{MLFN} 			& 81.2 & - 	  & - 	 & 62.8 \\
					DuATM \cite{DuATM} 			& 81.8 & 90.2 & 95.4 & 64.6 \\
					\hline
					IDE \cite{IDE} 				& 72.3 & 86.2 & 89.5 & 51.8 \\
					PCB \cite{PCB} 				& 83.8 & 91.7 & 94.4 & 69.4 \\
					st-ReID \cite{st-ReID} 		& 94.5 & 96.8 & 97.1 & 92.7 \\
					\hline
					\textbf{IDE+UnityStyle} 	& 85.9 & 93.5 & 94.8 & 82.3 \\
					\textbf{PCB+UnityStyle} 	& 89.3 & 95.7 & 96.2 & 85.7 \\
					\textbf{st-ReID+UnityStyle} & \textbf{95.1} & \textbf{97.0} & \textbf{97.3} & \textbf{93.6} \\
					\hline
			\end{tabular}}
		\end{center}
		\caption{Evaluation on the DuckMTMC-ReID dataset.}
		\label{tab:duck}
	\end{table}

	\subsection{Experiment Analysis}
	
	In this section, we will determine the effectiveness of each proposed module, and whether UnityGAN and UnityStyle improve the results from the perspective of datasets.
	
	\begin{figure}[t]
		\vspace{-0.4 em}
		\begin{center}
			\includegraphics[width=1\linewidth]{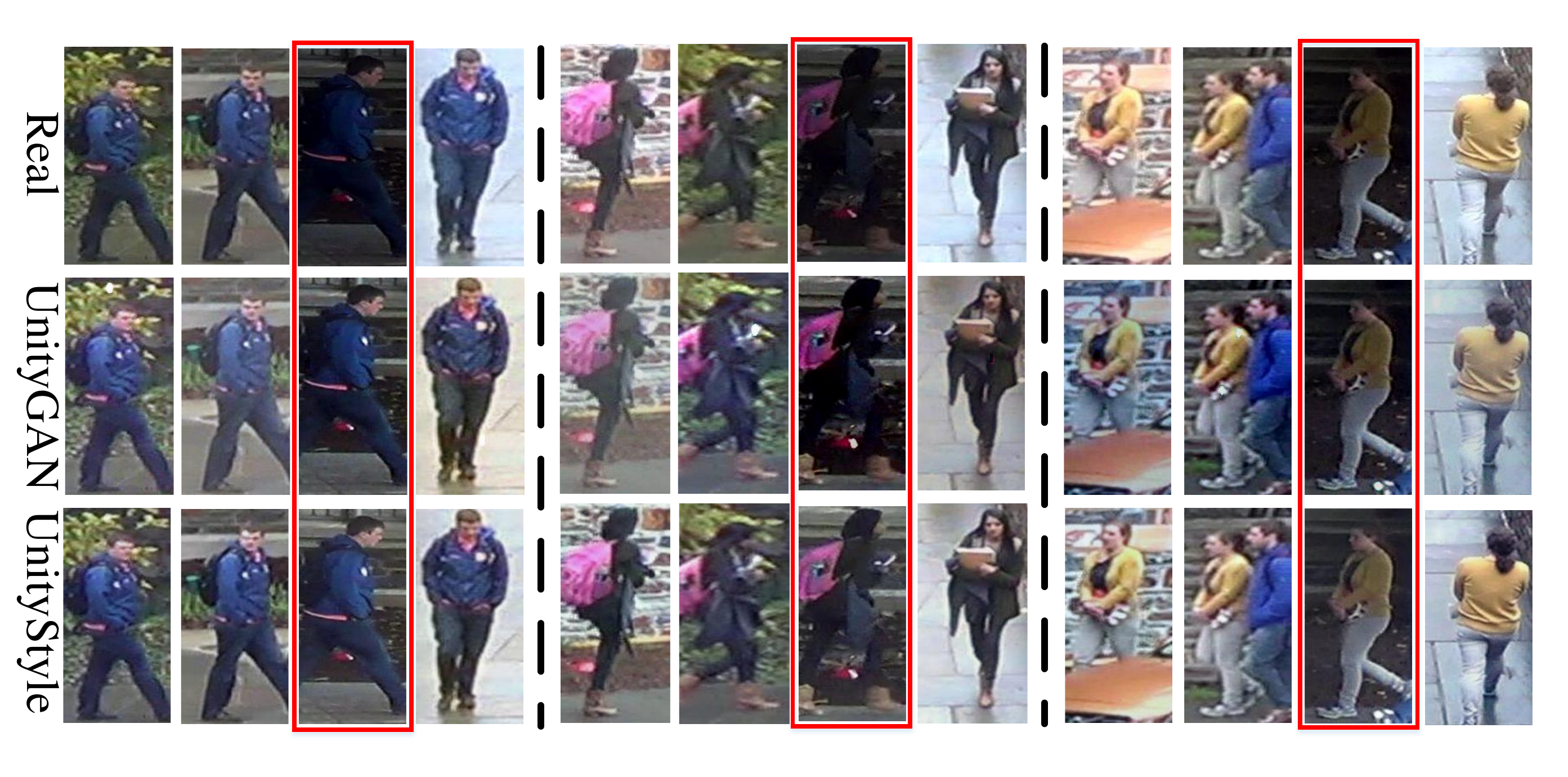}
		\end{center}
		\vspace{-1.4 em}
		\caption{Some samples in DuckMTMC-ReID. Real images, Images generated by UnityGAN without style attention module and UnityStyle images. With the Style Attention module, our model can generate style-stabilized images. We can clearly observe that the dark image (Red box) been enhanced in appearance, and the image generated by UnityStyle has a uniform style and lighting.}
		\label{fig:UnityStyle}
	\end{figure}
	
	\begin{figure}
		\vspace{-0.4 em}
		\begin{center}
			\includegraphics[width=1\linewidth]{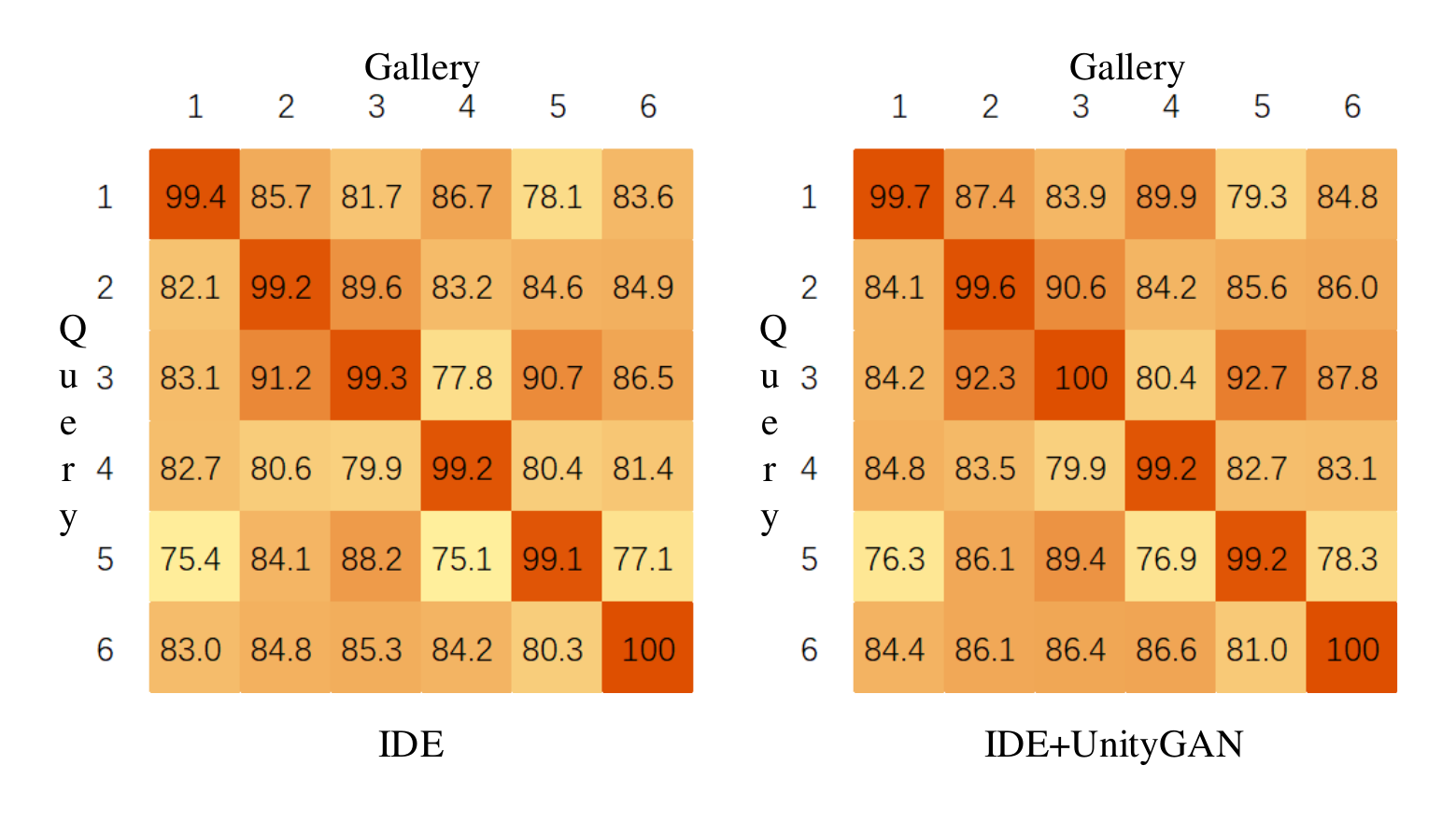}
		\end{center}
		\vspace{-1.4 em}
		\caption{Accuracy comparison between \textbf{IDE} and \textbf{IDE+UnityGAN} with different cameras. Different gallery cameras can be seen with a horizontal orientation and different query cameras with a vertical orientation.}
		\label{fig:IBNTEST}
	\end{figure}
	
	\subsubsection{UnityGAN}
	We contrast the test performance of IBN and IBN+UnityGAN. To clearly show the test result changes between the same camera and different cameras, we compare the accuracy of each camera's query with the camera's gallery separately. In Fig. \ref{fig:IBNTEST}, UnityGAN has a positive effect on IDE, although UnityGAN generates unstable style images (Fig. \ref{fig:UnityStyle}).
	
	\begin{figure}
		\vspace{-0.4 em}
		\begin{center}
			\includegraphics[width=1\linewidth]{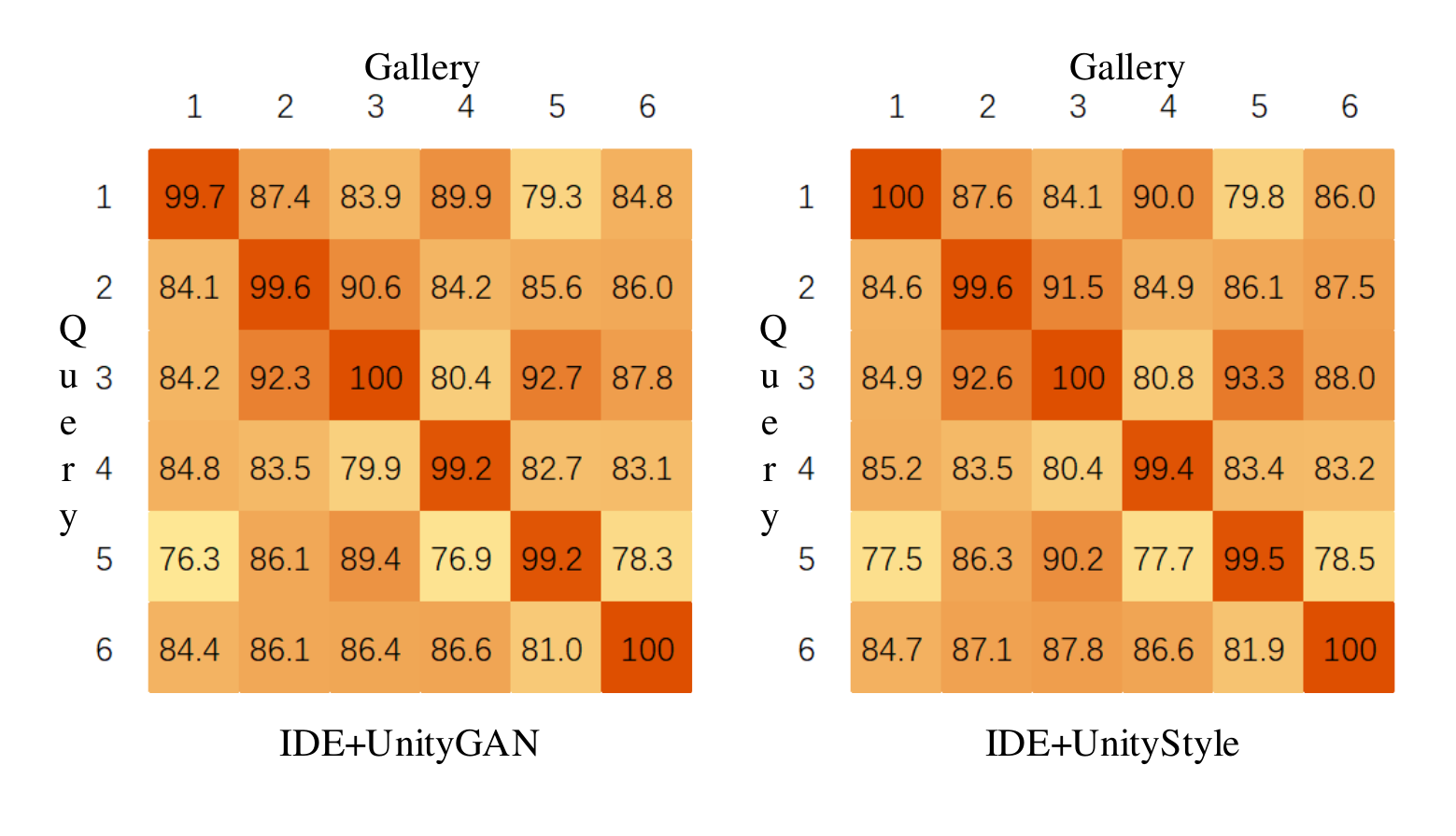}
		\end{center}
		\vspace{-1.4 em}
		\caption{Accuracy comparison between \textbf{IDE+UnityGAN} and \textbf{IDE+UnityStyle} with different cameras. Different gallery cameras can be seen with a horizontal orientation and different query cameras with a vertical orientation.}
		\label{fig:USTEST}
	\end{figure}
	
	\begin{figure}
		\vspace{-0.4 em}
		\begin{center}
			\includegraphics[width=1\linewidth]{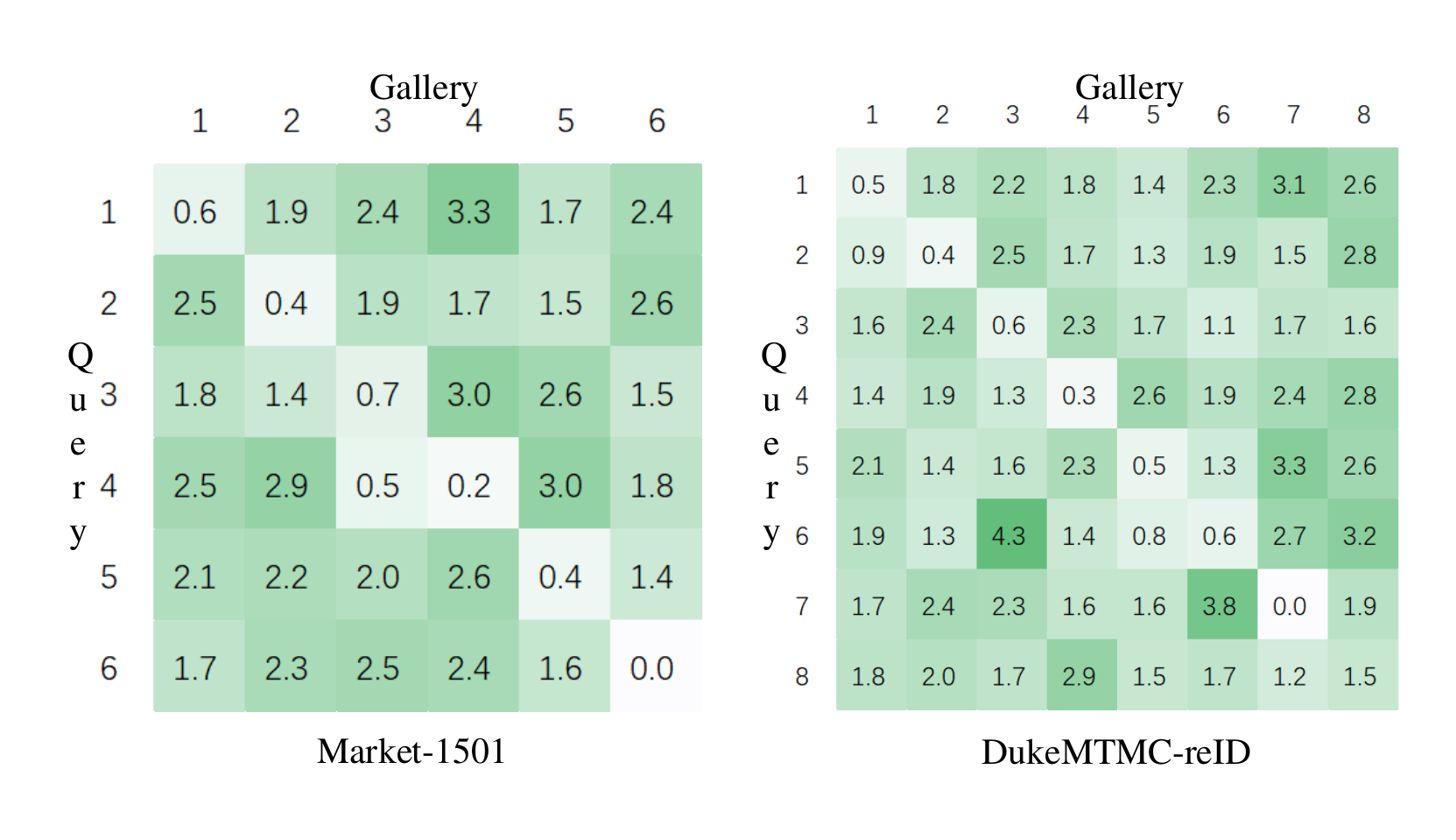}
		\end{center}
		\vspace{-1.4 em}
		\caption{The image shows the comparison of results before and after using \textbf{UnityStyle} under two data sets.}
		\label{fig:IBNvsUS}
	\end{figure}
	
	\subsubsection{UnityStyle}
	We propose the Style Attention Module and introduce the concept of UnityStyle, and we hope to use UnityStyle to remove the style differences of different camera images as much as possible to improve the results. In this section, we compare the test results of whether we use the UnityStyle. Fig. \ref{fig:UnityStyle} visually shows the impact of the Style Attention Module. DuckMTMC-ReID has greater camera views changes than Market-1501, and our method can handle these changes. To clearly show the changes in the test results of the same and different cameras, we compare the accuracy between each camera's query and each camera's gallery separately. As can be seen from Fig. \ref{fig:USTEST}, after adding UnityStyle, the accuracy of different cameras is further increased, and the accuracy of the same cameras has also risen even though it has reached a very high level. It shows that the introduction of UnityStyle can improve the adaptability to style changes. Fig. \ref{fig:IBNvsUS} can clearly show the promotion by IDE after using UnityStyle. 
	
	\subsubsection{Ablation Study}
	Table \ref{tab:IBN-UNI} shows the effect of using each module separately on the results of the two data sets. The IDE model is the baseline method, the UnityGAN model uses the unstable-style image to enhance the result in train phase, the UnityStyle model uses the UnityStyle image to enhance the result in train and test phase, and the UnityStyle+RE introduces re-ranking technology based on the UnityStyle model. From the results, we can see that the results are constantly improving as the introduction of the model. It's worth noting that we can achieve more significant improvement by using re-ranking technology after introducing UnityStyle, which means that UnityStyle increases the probability that there is a correct answer in the candidate, and re-ranking technology further increases the probability of top-1 the answer is selected. The experimental results prove that the combination of each part can achieve excellent results, and each component of the proposed method is mutually inseparable.
	
	\begin{table}
		\small
		\begin{center}
			\begin{tabular}{|l|c c|c c|}
				\hline
				Dataset&\multicolumn{2}{|c|}{Market-1501}&\multicolumn{2}{|c|}{DuckMTMC-ReID} \\
				\hline
				\hline
				Measure (\%) & top-1 & mAP & top-1 & mAP\\
				\hline
				IDE & 85.7 & 65.9 & 72.3 & 51.8\\
				IDE+RE & 87.0 & 77.5 & 73.5 & 69.4\\
				\hline
				UnityGAN & 89.8 & 73.0 & 79.1 & 60.6\\
				UnityGAN+RE & 90.9 & 83.1 & 82.7 & 76.5\\
				\hline
				\textbf{UnityStyle} & 91.8 & 76.5 & 82.1 & 65.2\\
				\textbf{UnityStyle+RE} & \textbf{93.2} & \textbf{89.3} & \textbf{85.9} & \textbf{82.3}\\
				\hline
			\end{tabular}
		\end{center}
		\caption{The effect of each module. With the addition of modules, the model's effect is getting better and better. RE: re-ranking.}
		\label{tab:IBN-UNI}
	\end{table}
	
	\section{Conclusion}
	
	In this paper, we propose UnityStyle method, which can smooth the style disparities within the same camera and across different cameras. We firstly create UnityGAN to learn the style changes between cameras, producing shape-stable style-unity images for each camera. Motivated by the fact that structural information is contained in shallow layers, we add skip connections between multi-depth layers, thus retain more structural information and make the generated image structure more stable. Then, we propose UnityStyle images to eliminate style differences between different images, which makes a better match between query and gallery. It is ensured that images with a large difference in unity-style have a small effect on the style of the generated image, and therefore we can generate images with unity-style.These advantages make the proposed method perform better than existing methods.
	
	\section*{Acknowledgments}
	
	This work is supported in part by China National 973 program 2014CB340301 and NSFC grant 6197023605, 61973250 and 61702415, Australian Research Council (ARC) Discovery Early Career Researcher Award (DECRA) under grant no. DE190100626, Air Force Research Laboratory and DARPA under agreement number FA8750-19-2-0501.

	{\small
		\bibliographystyle{ieee_fullname}
		\bibliography{egbib}
	}
	
\end{document}